\documentclass{clv3}
\usepackage[table]{xcolor}
\definecolor{darkblue}{rgb}{0, 0, 0.5}

\usepackage[T1]{fontenc}

\usepackage[utf8]{inputenc}

\usepackage{microtype}
\usepackage{xspace,mfirstuc,tabulary}
\usepackage[algo2e,linesnumbered,ruled,vlined]{algorithm2e}
\usepackage{enumitem}

\usepackage{pgfplots}
\usepackage{makecell}
\usepackage{multicol}
\usepackage{subfig}
\usepackage{amsthm}
\usepackage{amsmath}
\usepackage{mathtools}
\usepackage{amssymb}
\usepackage{multirow}
\usepackage{comment}
\usepackage{array}
\usepackage{tabularx}
\definecolor{darkspringgreen}{rgb}{0.09, 0.45, 0.27}
\usepackage{graphicx}
\usepackage{tikz}
\usepackage[singlelinecheck=false,justification=centering]{caption}
\usepackage{subcaption}

\usepackage{scalerel,stackengine}
\stackMath
\newcommand\reallywidehat[1]{%
\savestack{\tmpbox}{\stretchto{%
  \scaleto{%
    \scalerel*[\widthof{\ensuremath{#1}}]{\kern-.6pt\bigwedge\kern-.6pt}%
    {\rule[-\textheight/2]{1ex}{\textheight}}
  }{\textheight}%
}{0.5ex}}%
\stackon[1pt]{#1}{\tmpbox}%
}
\makeatletter
\renewcommand\@makefntext[1]{\leftskip=2em\hskip-2em\@makefnmark#1}
\makeatother

\usepackage{algorithm,algpseudocode}
\newlength{\bibparskip}
\let\oldthebibliography\thebibliography
\renewcommand\thebibliography[1]{%
  \oldthebibliography{#1}%
  \setlength{\parskip}{\bibitemsep}%
  \setlength{\itemsep}{\bibparskip}%
}

\usepackage{enumitem}
\setlist{nolistsep}
\newcolumntype{P}[1]{>{\centering\arraybackslash}p{#1}}

\renewcommand{\appendixsection}[1]{
  \section{Appendix \Alph{section}: #1}%
}
\renewcommand\appendix{%
   \setcounter{section}{0}
   \setcounter{figure}{0}
   \setcounter{table}{0}
   \setcounter{equation}{0}
   \renewcommand{\thesection}{\Alph{section}}
   \renewcommand{\theequation}{\Alph{section}.\arabic{equation}}
}

\definecolor{mycolor}{RGB}{202, 42, 219}

\newcommand{\attack}{\textsc{IDT}\xspace}

\usepackage{booktabs}
\usepackage{caption}

\usepackage{supertabular}
\usepackage{adjustbox}
\usepackage{url}
\usepackage{hyperref}
\usepackage{cleveref}
\hypersetup{colorlinks=true,citecolor=darkblue, linkcolor=darkblue, urlcolor=darkblue}

\begin{document}


\runningtitle{Adversarial Attacks for Privacy Protection}
\runningauthor{Faustini, Tonni, McIver, Dras}



\title{IDT: Dual-Task Adversarial Attacks for Privacy Protection}



\author{Pedro Faustini}
\affil{School of Computing, Macquarie University\\pedro.arrudafaustini@hdr.mq.edu.au}
\author{Shakila Mahjabin Tonni}
\affil{School of Computing, Macquarie University\\shakila.tonni@mq.edu.au}
\author{Annabelle McIver}
\affil{School of Computing, Macquarie University\\annabelle.mciver@mq.edu.au}
\author{Qiongkai Xu}
\affil{School of Computing, Macquarie University\\qiongkai.xu@mq.edu.au}
\author{Mark Dras}
\affil{School of Computing, Macquarie University\\mark.dras@mq.edu.au}


\maketitle
\begin{abstract}

Natural language processing (NLP) models may leak private information in different ways, including membership inference, reconstruction or attribute inference attacks. Sensitive information may not be explicit in the text, but hidden in underlying writing characteristics.
Methods to protect privacy can involve using representations inside models that are demonstrated not to detect sensitive attributes or --- for instance, in cases where users might not trust a model, the sort of scenario of interest here --- changing the \emph{raw} text before models can have access to it. 
The goal is to rewrite text to prevent someone from inferring a sensitive attribute (e.g. the gender of the author, or their location by the writing style) whilst keeping the text useful for its original intention (e.g. the sentiment of a product review). 
The few works tackling this have focused on generative techniques. However, these often create extensively different texts from the original ones or face problems such as mode collapse.
This paper explores a novel adaptation of adversarial attack techniques to manipulate a text to deceive a classifier w.r.t one task (\textbf{privacy}) whilst keeping the predictions of another classifier trained for another task (\textbf{utility}) unchanged.  
We propose \attack, a method that analyses predictions made by auxiliary and interpretable models to identify which tokens are important to change for the privacy task, and which ones should be kept for the utility task. 
We evaluate different datasets for NLP suitable for different tasks. Automatic and human evaluations show that IDT retains the utility of text, while also outperforming existing methods when deceiving a classifier w.r.t privacy task.  
\end{abstract}





\section{Introduction}\label{sec:1_introduction}
There are many different ways that NLP models can be vulnerable to leaking private information, and many ways that the developers or users of NLP systems can attempt to preserve privacy \citep{sousa-kern:2023:AIR}.  
Much of the focus has been on protecting the training set to be resistant to e.g. membership inference or reconstruction; such work has often used a framework with mathematical guarantees like central Differential Privacy (DP), for instance via training techniques like DP-SGD \citep{abadi-etal:2016:CCS}.
Privacy violations via \textit{inference} are another class of attack: while the ability to infer potentially sensitive information (say, age or gender) has long been known \citep[for example]{schler-etal:2006}, there has been recent concern at the demonstrated ease of this kind of inference via large language models \citep{staab2024beyond:ICLR}. 
 The sort of scenario of interest here might include an individual who posts a review or comments on a website or social media forum like Reddit, but does not wish a sensitive attribute that may enable identification to be discoverable; the same techniques that facilitate automatically determining, for example, aggregate sentiment, can also automatically find traces of these sensitive attributes.


While there is a broad class of privacy protections against this kind of inference that involve making a model trustworthy, and learning latent representations that while still useful for the primary or utility task (say, sentiment classification) do not leak sensitive information \citep[for example]{li-etal-2018-towards,yu-etal:2022:ICLR:differentially}, we are interested in the kind of scenario where the model does not provide this protection or where the individual may not trust the model.

Another kind of protection, the one of interest in this paper, is in modifying \textit{raw text} before it is sent to a model.  
In NLP, works have largely focused on generative approaches that aim to provide (local) DP guarantees of a general sort \citep{krishna-etal-2021-adept,weggenmann-etal:2022:WWW,chen-etal-2023-customized}.  However, these often result in texts that differ extensively from the original one and typically do not have a specific sensitive attribute they aim to protect.
%
%
%
%
%
A small number of other works have aimed to change the text in a way that does protect some specific attribute. \citet{xu-etal-2019-privacy} proposed an adversarial trainer inspired in Generative Adversarial Networks \citep{NIPS2014_5ca3e9b1} in which a linear classifier sends a signal to a Transformer \cite{10.5555/3295222.3295349} to paraphrase a text towards a different class. Another approach, from \citet{tokpo-calders-2022-text}, trains a model to replace tokens with similar ones, but it needs to adopt a soft sampling to allow gradients to backpropagate. Both approaches, while potentially producing natural-looking text that achieves the goal, face challenges typical of generative models such as mode collapse, which is the case when mostly non-diverse samples are generated \citep{huijben-etal:2023:TPAMI}.

In this paper, then, we have a similar goal: a method that only minimally rewrites a text from an author's intended original, maintaining utility while empirically preserving attribute privacy in the sense of \citet{coavoux-etal:2018:EMNLP:privacy} or \citet{li-etal-2018-towards}.  Our aim is to define a method that is robust and effective across datasets and tasks and that does not suffer from issues like mode collapse.
To do this, we draw on the notion of adversarial attacks in a novel way.  Adversarial attacks fool classification models by making subtle changes to their inputs. While they have been widely studied in the context of classification tasks, they have not been applied in scenarios we refer to as \textit{dual-task}, where a piece of text is associated with labels for two distinct classification tasks. 
In the kinds of scenario of interest here, we apply them to fool one kind of classifier (for a sensitive attribute) while retaining utility on another classifier.


We thus propose \textbf{I}nterpretable \textbf{D}ual-\textbf{T}ask --- \textbf{\attack}. 
In addition to adversarial attacks, we take inspiration from membership inference attacks \citep{7958568} and train \emph{auxiliary models}, which mimic the target models. We analyse the predictions made by them with an interpretable model 
that ranks words according to their importance for a given classification task. This way, we can find important words for the privacy and the utility tasks. We then modify relevant tokens for the private task, but keep the important ones for the utility task unchanged.

Our contributions are as follows:


\begin{itemize}
    \item We design a method based on adversarial attacks for rewriting text, IDT, such that the rewritten text preserves performance on some utility task while avoiding detection of some potentially sensitive attribute.

    \item We carry out an extensive evaluation of IDT with other types of text rewriting for privacy purposes.  We show that IDT in general outperforms these other approaches, often by a large margin; moreover, we also show that some classes of empirical privacy protection methods do not protect in our kind of scenario.
\end{itemize}

\section{Related work}\label{sec:related_work}

In the following, we briefly review some relevant NLP work on privacy, followed by work on adversarial attacks, on which our method is based. These subsequent works lie on single classification problems, and often in binary classification tasks. Therefore, we identify a gap in the literature concerning the privacy-utility trade-off in multi-class problems, as well as cases where instances may belong to several classes.



\subsection{Privacy in NLP}

Previous works usually either ensured privacy by modifying the embedding space that represents the texts, or by altering the raw strings with some obfuscation technique, such as differential privacy. \citet{sousa-kern:2023:AIR} give a detailed survey of these; we note below some particular instances.

Regarding the first approach, for instance, \citet{10.1007/978-3-030-17138-4_6}, \citet{plant-etal-2021-cape} and  \citet{meehan-etal-2022-sentence} apply different types of differential privacy noise to various embedding representations in order to prevent sensitive information from being inferred from the data. Aiming to protect privacy via an empirical demonstration rather than using the mathematical guarantees of differential privacy, on the other hand, \citet{li-etal-2018-towards} trained a generative model in an adversarial fashion in which the model generates a (latent) representation \textbf{h} of a text for some utility task (e.g. POS tagging). At the same time, \textbf{h} is designed to be a bad representation for sensitive attributes (such as age or gender). 
\citet{coavoux-etal:2018:EMNLP:privacy} provide another approach to the same task of producing representations that aim to avoid encoding sensitive attributes.

These kinds of approaches are suitable for scenarios where, for example, a system should employ representations that fulfill its primary functionality well but where the system creator wants to convince users that the system's decisions will not take account of their sensitive attributes.  Another perspective --- from the point of view of a user who may not trust a system creator --- is to change the raw input before it is passed to a system.


Works from this other perspective, usually within a local differential privacy framework, create alternative texts by adding noise to an encoder or decoder \citep{igamberdiev-etal-2022-dp, yue-etal-2021-differential}, and usually evaluate privacy w.r.t. the tokens being replaced, rather than a particular (labelled) task. For example, CusText \citep{chen-etal-2023-customized} and DP-Prompt \citep{utpala-etal-2023-locally} replace tokens according to an obfuscated list of semantic similarities, while the approach of \citet{weggenmann-etal:2022:WWW} perturbs the latent vectors and later decodes them into text.  The texts produced by these approaches often differ greatly from the original (e.g. METEOR similarity for \citet{weggenmann-etal:2022:WWW} was very low, 5\% or 9\% according to the dataset, indicating that many tokens were replaced). Further, it is a more general notion of protection provided here, with \citet{igamberdiev-habernal-2023-dp} emphasising the need to be clear on what exactly is being privatised. For instance, paraphrasing has been studied by \citet{ponomareva-etal-2022-training} by pre-training a T5 model \citep{2020t5} and its tokenizer with DP to prevent training data from leaking when the model generates text. In another example, CusText, evaluated privacy under an unlabelled task of predicting the original tokens using a BERT-MLM model against the sanitised sentence. In their survey, \citet{klymenko-etal:2022:differential} discuss that practical benefits of DP in NLP applications lie in the individual space: an attacker could still infer tokens from text, but there would be some uncertainty as to whether such token was indeed the original one. However, it is unclear how to control the effect of sanitisation for a specific task of concern in our sort of attribute inference scenario under such a framework.

Also, it is common for the $\epsilon$ parameter from DP to be large, in the order of hundreds or even thousands, which means low privacy guarantees. \citet{igamberdiev-habernal-2023-dp} acknowledge that the lowest useful $\epsilon$ is too high for real-world applications, given that different applications in the literature range $\epsilon$ from 0.01 to 10. \citet{sousa-kern:2023:AIR} remark that privacy in NLP is exchanged for performance: for example, adding noise to the embeddings may cause semantic disruption, compromising the final results of downstream tasks. To avoid that, in DP-SGD works, such as from \citet{kerrigan-etal-2020-differentially}, non private datasets are still necessary for pretraining models, and the private data is used in later stages, for fine-tuning. 

A related field is text sanitisation, in which a model rewrites the input by removing or replacing personally identifiable information (PII) \citep{albanese2023text}. The difference is that the property to keep private is not a task, but individual tokens. As a consequence, the narrow vocabulary that can be changed, coupled with the limited options for the replacements (usually a predefined token), may not be enough to obfuscate high-level properties one may wish to hide.

The works that have the same goal as ours --- that is, to rewrite raw text input with the goal of concealing a sensitive attribute from detection --- come from \citet{xu-etal-2019-privacy} and \citet{tokpo-calders-2022-text}. The first explored how back-translation reduces the leakage of sensitive information. Their approach is heavily inspired by GANs, and thus sustaining training stability to avoid mode collapse is difficult. 
In terms of evaluation, it was assessed on three datasets were designed for only binary classification problems, and the utility task was restricted to sentiment analysis; we are interested in evaluating contexts beyond this.
A similar issue with potential mode collapse happens with the work in progress presented in \citet{tokpo-calders-2022-text}. It changes the style of a text w.r.t a task by replacing individual tokens. However, their method requires a soft sampling, and finding the appropriate $\tau$ temperature parameter is challenging.  In contrast to these, we use an optimisation-based adversarial attack approach, that can be expected to produce rewritten texts in a more reliable fashion. 


\subsection{Adversarial Attacks}\label{sec:rl_attacks}

Adversarial attacks are methods that strategically modify input text to fool a model into making an incorrect prediction.
There are now a number of adversarial attacks against text in the literature: in terms of overviews, \citet{zhang-etal:2020} and \citet{qiu-etal:2022:Neurocomputing} provide surveys, while \citet{dyrmishi-etal-2023-humans} empirically compares some of the major methods.
\citet{morris-etal-2020-textattack}, in designing a common framework for adversarial attacks called TextAttack,\footnote{\url{https://github.com/QData/TextAttack}} 
showed how previously disparate adversarial attack methods could be thought of in a unified way, consisting of four types of components: a goal function, operationally specifying the method goal such as changing classifier prediction or changing words used in translation; a set of constraints reflecting desiderata for adversarial text quality, such as semantic similarity to the original text or grammaticality; transformations, the set of allowable perturbation types; and search methods, the approach (e.g. genetic algorithm) to finding an adversarial text using the transformations that satisfies the constraints and the goal function. Adversarial attacks are shown transferable from shadow models to the original models~\citep{he-etal-2021-model}.
We discuss below the three specific adversarial attacks that we use as the basis for our privacy-protecting method; all three are implemented in the TextAttack framework.

TextBugger \citep{DBLP:conf/ndss/LiJDLW19} works by comparing the prediction before
and after removing a word or character to measure its influence. The utility is measured in terms of four similarity metrics between the sentences, which should be above a specific threshold. Moreover, the authors carried out a human evaluation which concluded that the sentences retained their sentiment/toxicity.

Another popular attack is TextFooler, by \citet{Jin_Jin_Zhou_Szolovits_2020}. It replaces words according to a list of constraints as cosine similarity and maintains POS tags\footnote{The TextAttack implementation relaxes this constraint by allowing nouns to be swapped by verbs and vice-versa.}. However, their evaluation compared the accuracy scores of original and adversarial sentences for one task per dataset. Utility was measured by reporting the ratio of perturbed words and the semantic similarity between the texts.

\citet{garg-ramakrishnan-2020-bae} proposed BAE, which masks tokens and uses the BERT-MLM to create alternatives to replace them. Utility, as in the previous works, is assessed by calculating semantic similarity, with the addition of humans evaluating how suspicious the sentences were altered by a machine and how they rate them w.r.t their sentiment --- which is the same task used for the attacks.



\section{Problem definition}\label{sec:definition}




We define our problem as a dual-task optimisation problem problem on two tasks for privacy ($p$) and utility ($u$). 
Let $D = \{\mathcal{X}, \mathcal{Y}^p, \mathcal{Y}^u \}$ be a dataset where $\mathcal{X}$ is a collection of texts, and $\mathcal{Y}^p$ and $\mathcal{Y}^u$ are their labels for task $p$ (privacy) and task $u$ (utility). A classifier $f_p$ assigns instances $x \in \mathcal{X}$ to a label $y_p \in \mathcal{Y}^p$, whereas $f_u$ assigns a label $y_u \in \mathcal{Y}^u$ to instances $x \in \mathcal{X}$. An attacker aims at generating a perturbed text $x'$ from a genuine text $x \in \mathcal{X}$. 

An \textit{successful} attack is defined as $f_{p}(x) \neq f_{p}(x')  $  and $f_{u}(x) = f_{u}(x')$: the utility of $u$ for $x'$ is kept while the privacy of $p$ is retained by forcing $f_{p}$ to misclassify $x'$.

Intuitively, our problem deals with making minimal editing to texts so that their usefulness remains not merely in terms of how close the attacked sentences are to their original counterparts, but for a utility task. For instance, posts in marketplaces are \textit{useful} if they keep their sentiment towards the products, whereas potentially sensitive identifiable information w.r.t. the writer (e.g. their gender, or age) can be hidden.
We are principally interested in scenarios where there are machine learning models that can detect the sensitive attribute, and the user might want to avoid such detection.%
\footnote{A real-world instance of this was the implicit detection of gender in recruitment processes using machine learning: \url{https://bit.ly/2ycdnVV}.}

\begin{figure}
        \centering
        \includegraphics[scale=0.5]{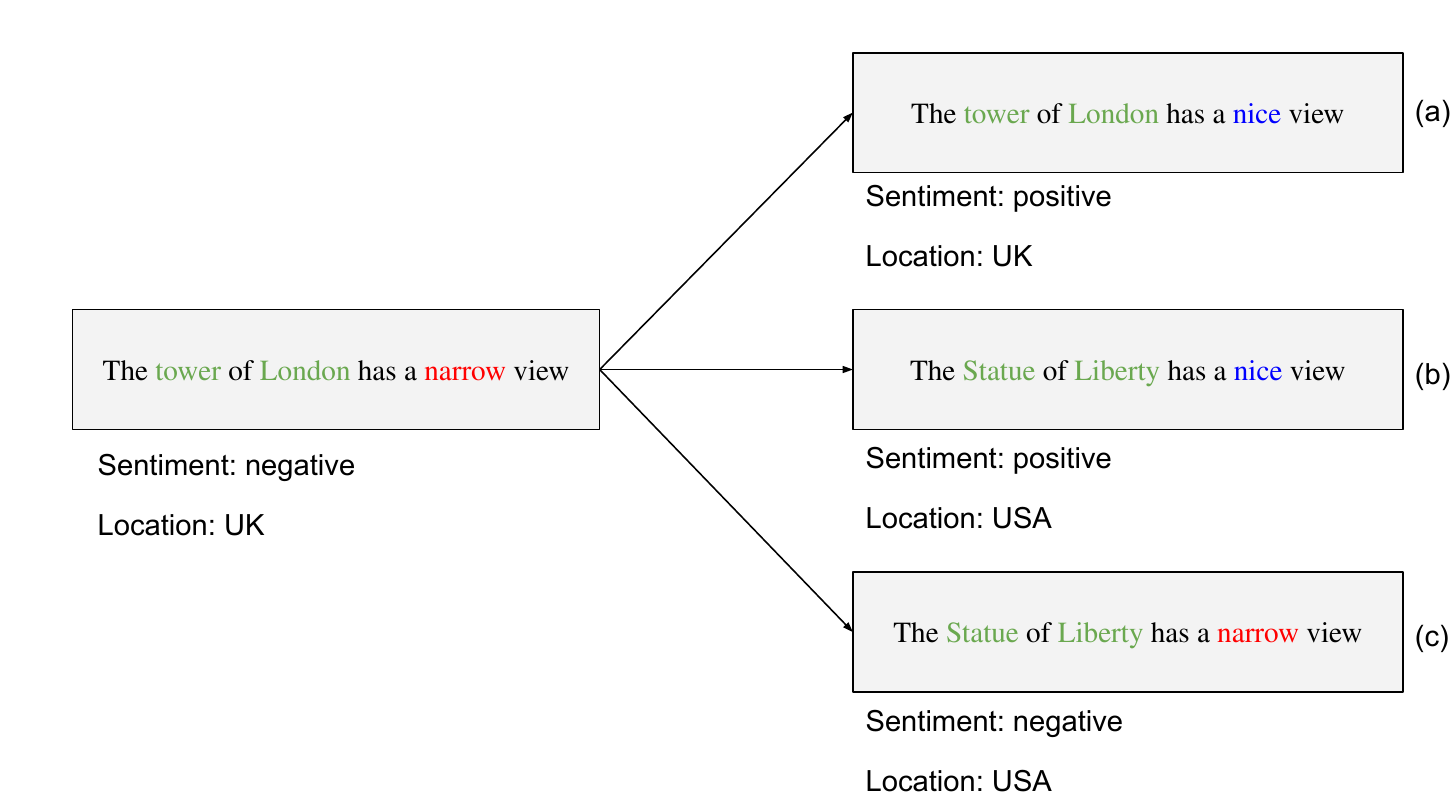}
        \caption{By identifying which words are important to each classification task, one can rewrite the sentence aiming to deceive a classifier w.r.t. a particular task only.}
        \label{fig:flow}
        \end{figure}

We show in Figure~\ref{fig:flow} a toy example depicting how a single sentence can be associated to different tasks (sentiment as $p$ and topic as $u$), and how changing particular words may affect the prediction for each task. Without looking at any particular class, a sentence may be rewritten in a way to lose all its semantic content (as in rewriting (b)). This can be avoided by defining one task to change the prediction of another to keep it.

Note that while the novel aspects of our work primarily concern the adaptation of adversarial attack methods to achieve this goal, our problem formulation above does differ in some respects from prior work.  This is because our notions of utility and privacy are defined with respect to some concrete (classification) tasks, with prior work differences in focus and framing inferrable from what they evaluate.
The closest comparison works, \citet{xu-etal-2019-privacy} and \citet{tokpo-calders-2022-text}, do measure privacy with respect to classification performance on a sensitive attribute.  However, \citet{xu-etal-2019-privacy}'s primary non-privacy evaluation is on ``linguistic quality'', including both automatic metrics (e.g. BLEU, GLEU and Word Mover Distance) and human evaluation of fluency and relevance, or semantic closeness to original (although a secondary analysis does also look at classification accuracy of a utility task). \citet{tokpo-calders-2022-text} likewise look at semantic closeness to original as measured by cosine similarity.  Other types of private rewriting within a DP framework, such as CusText \citep{chen-etal-2023-customized}, are not concerned with a specific sensitive attribute, as they typically evaluate the privacy protection only in terms of privacy budget $\epsilon$.

\section{\attack Attack}\label{sec:attack}

\attack modifies important tokens for the privacy task, whilst keeping important ones for the utility task. We take inspiration from membership inference attacks \citep{7958568} and train shadow (auxiliary) models, whose goal is to mimic the target models. Attackers, trained based on the outputs of shadow models, infer if an object was used for training by looking at the class probabilities outputs. In our case, our shadow models, hereafter called auxiliary models, are used to find the important words for each classification task.

Our attack has the following assumptions:

\begin{itemize}
    \item As well as with the membership inference attacks, we assume that the attacker has data to train auxiliary models. 
    \item The attacker uses the auxiliary models to create adversarial sentences against the victim.
    \item Previous works query the target several times to create the adversarial sentences; \attack creates them offline with auxiliary models. Thus, it only queries the target once per sentence.
\end{itemize}



Algorithm \ref{alg:attack} depicts its pseudocode. It has three main blocks; comments highlight each one.
    
\RestyleAlgo{ruled}
\SetKwComment{Comment}{/* }{ */}
\begin{algorithm}[t!]
\caption{\attack Attack}\label{alg:attack}
\KwData{Sentence $X = \{w_1, w_2, ...,w_n\}$, ground truths $y^{p, u}$, auxiliary models $\theta_{p, u}$, word embeddings $E$ over the vocabulary $V$, $k$ nearest words to consider, query budget $q$ }
\KwResult{List of adversarial $Xs'$}

\Comment{Searching for potential words in X to replace}

$explanations_{priv} \gets explain(\theta_p, X) $;
$explanations_{util} \gets explain(\theta_u, X) $;

$top\_words_{priv} \gets $ Get the top influential words from $explanations_{priv}$;

$top\_words_{util} \gets $ Get the top influential words from $explanations_{util}$;

final\_words $\gets \emptyset$;

\For{$w \in top\_words_{priv}$}{
    \If{not $w  \in top\_words_{util}$  }{
        $final\_words \gets final\_words \cup \{w\}$;
    }
}

\Comment{Searching for the most similar words for each word in final\_words}

nearest\_words $\gets \{\}$;

\For{$w \in final\_words$}{

nearest\_words\textsubscript{w} $\gets$ $k$ most similar words to $w \in V$ ;

}

\Comment{Searching for adversarial sentences by sampling i words each time}

Xs' $\gets \emptyset$ ;

\For{$i$ in $len(final\_words$)}{
    \Repeat{$q$ times}{

        words\_to\_change $\gets$ sample(final\_words, i);
        
        X' $\gets$ X;

        \For{w in words\_to\_change}{

            w' $\gets$ random(nearest\_words[token]);
    
            temp $\gets$ X'.replace(token, w');
    
            \If{POSTags(temp) == POSTags(X)}{
            
                X' $\gets$ temp;

                Xs' $\gets$ Xs' $\cup$ X';
            }
        }
    }

}
\end{algorithm}



Lines 1 to 8 select potential words to change, with the core of the choice of words happening between lines 5 and 7. 
Explanations\textsubscript{priv, util} are maps from a token to its score according to the interpretable model running through the auxiliary models $\theta_{priv}$ and $\theta_{util}$. 
In our experiments, we used the Layer Integrated Gradients from \citet{mudrakarta-etal-2018-model}, available in the Captum library \citep{DBLP:journals/corr/abs-2009-07896}. We emphasise that any interpretable method which assigns scores to tokens can be adopted. Then, the influential words (those assigned a positive value) for the privacy and utility tasks are selected, respectively. Then, only those important for the privacy task, but which do not appear in the utility list, are kept. The rationale is to modify sensitive words for privacy whilst preserving utility.

Lines 9 to 11 select the $k$ most similar words for each word in the sentence. In our experiments, we converted words to latent space with Glove embeddings \citep{pennington-etal-2014-glove}.

Last, lines 13 to 22 search for valid adversarial texts. Adversarial sentences are created by replacing words with similar ones in increasing amounts. We also set the constraint that adversarial texts must match the POS tags of the original one, as several previous works have shown this constraint to be effective \citep{Jin_Jin_Zhou_Szolovits_2020, garg-ramakrishnan-2020-bae, yoo-qi-2021-towards-improving}.

From the generated adversarial  Xs', we select the one that deceives the auxiliary model $\theta_{priv}$ with the highest confidence score. 
\section{Experimental setup}\label{sec:setup}

We used distilled RoBERTa \citep{DBLP:journals/corr/abs-1907-11692}, pretrained base models from the Transformers library \citep{wolf-etal-2020-transformers} as the architecture for both victim and auxiliary models. We also design a further analysis with differing architecture by using a distilled GPT2 \citep{radford2019language} as victim.

\paragraph{Datasets}
We pick datasets suitable for more than one task. We select one attribute as sensitive, for which the adversarial text must lead to a different classification than the original, and another for utility, for which the classification should remain the same for original and adversarial.

\textbf{TrustPilot} \citep{hovy-etal:2015:WWW} is a core dataset for investigating privacy, and it has been adopted by many works studying attribute inference attacks \citep{he-etal-2022-extracted, coavoux-etal:2018:EMNLP:privacy, li-etal-2018-towards}. It contains reviews alongside attributes such as numerical rating, gender, location, and year. We use the ratings as the utility task, converting the 1-5 scale into bad, mixed, and good categories. For the sensitive attribute, we use gender, age and location.

We follow previous works and treat age and gender as binary values. We split the age class between those born before 1967 and after 1977, leaving a 10-year gap in between. A similar approach was used by \citet{li-etal-2018-towards}. For location, we also follow  \citet{li-etal-2018-towards} and retain English reviews according to the Langid tool \citep{lui-baldwin-2012-langid} and ensure the texts are balanced amongst the five classes (Denmark, German, France, UK, US).

To evaluate generalisability across other attributes, we also experiment with two other datasets.
\textbf{TOEFL11} \citep{TOEFL11} was collected for the task of native language identification (NLI).\footnote{Not to be confused with Natural Language Inference, which is a different NLP task.} Each essay was written in English by learners from 11 other languages. Each document is also labelled by one of the 8 topics the essay is about. We set the native language as the sensitive attribute, the topic for utility, and split the essays into sentences. Thus, we evaluate privacy and utility under \textit{multi-class tasks}.

\textbf{Shakespeare} \citep{xu2012paraphrasing} contains sentences from 17 plays, labelled according to their writing style (modern or old). We use the style as the attribute to change and the play as the one to preserve the classification. We note that utility retention should be harder on this dataset, given its large number of classes.

We split each dataset into target and auxiliary model data. Specific details about the splits can be found in Table \ref{tab:dataset_splits} in the Appendix.

For each dataset, we retrieved the texts that were correctly classified for both utility and privacy tasks. They are then passed through the auxiliary models, and adversarial texts are generated based on their explanations. The adversarials that deceived the auxiliary models in the privacy task, but not for utility, are queried against the victim. Table \ref{tab:dataset_tasks} summarises the privacy and utility tasks for each dataset, alongside the number of classes per task.

\setlength{\tabcolsep}{3pt}
\begin{table}[h]
    \centering
    \small
    \begin{tabular}{ccc|cc}
    \hline
        \textbf{Dataset} & \textbf{Privacy}  & \textbf{\#Classes}  & \textbf{Utility}  & \textbf{\#Classes} \\\hline
        
        TrustPilot\textsubscript{L} & Location & 5 & Rating & 3\\
        TrustPilot\textsubscript{G} & Gender & 2 & Rating & 3\\
        TrustPilot\textsubscript{A} & Age & 2 & Rating & 3\\
        
        TOEFL11 & NLI & 11 & Topic & 8\\
        Shakespeare & Style  & 2 & Play &17\\\hline
    \end{tabular}
    \caption{Privacy and utility tasks per dataset with the number of classes for each one.}
    \label{tab:dataset_tasks}
\end{table}



\paragraph{Baselines}\label{par:baselines}
To the best of our knowledge, there is no system that tackles our specific problem as we have framed it. Therefore, we adopt the back-translation model from \citet{xu-etal-2019-privacy} since it is the closest proposal to ours, as highlighted in Section \ref{sec:rl_attacks}. More details about its implementation can be found in Appendix \ref{app:backtranslation}.  We also considered the system of \citet{tokpo-calders-2022-text}.  However, we could not successfully train this on our datasets and tasks without mode collapse,\footnote{For instance, the sentence ``draw the curtains, just like that.'' collapses to ``draw the sword , barlow barlow barlow''} so we do not present results for it.

For this principal baseline, we re-implemented an \textbf{adversarial back-translator} that follows the proposal of \citet{xu-etal-2019-privacy}. Since the source code is not available, we made the following changes: the parallel corpora were generated with the pretrained MarianMT models (discussed following). The back-translator is a Bart model instead of the original (now outdated) Transformer \citep{10.5555/3295222.3295349}. The adversary is a BART classification head. Following their experiments, the $\alpha$ parameter was set to 1.0. More details are in Appendix \ref{app:backtranslation}.

Related to this, we evaluate how ordinary (non-adversarial) \textbf{back-translation} disrupts the texts, an approach that has been tried and used as a baseline by other works \citep[for example]{prabhumoye-etal-2018-style} including our primary baseline of \citet{xu-etal-2019-privacy}. Our back-translation baselines are two MarianMT pretrained models released by \citet{TiedemannThottingal:EAMT2020}. They were trained with the Opus corpus \citep{Tiedemann2009NewsFO}. One model translates sentences from English to French, and the other converts French texts back to English. The authors report BLEU scores ranging from 27.5 to 50.5 across 10 test sets for the first translator,\footnote{\url{https://huggingface.co/Helsinki-NLP/opus-mt-en-fr}} and from 26.2 to 57.5 for the second.\footnote{\url{https://huggingface.co/Helsinki-NLP/opus-mt-fr-en}}

For further baselines, we investigate whether solutions to related problems already solve ours; specifically, existing adversarial attacks and text sanitisation techniques.
We compare \attack to adversarial algorithms provided by TextAttack \citep{morris-etal-2020-textattack}, under different query budgets. High budgets are akin to a brute force search and can be prohibitively costly. By default, TextAttack queries the target model several times, which in reality may not be feasible due to defence mechanisms employed by servers. Thus, we also experiment when TextAttack creates adversarial texts against our auxiliary models and then they are tested against the target auxiliary, akin to our methodology for \attack.

We use \textbf{TextBugger} \citep{DBLP:conf/ndss/LiJDLW19}, \textbf{TextFooler} \citep{Jin_Jin_Zhou_Szolovits_2020} and \textbf{BAE} \citep{garg-ramakrishnan-2020-bae} as extra baselines. These algorithms were chosen because they alter texts in different levels of granularity: TextFooler replaces words based on similarity constraints, whereas BAE uses a BERT-MLM model to generate replacements for masked tokens. TextBugger performs character-level replacements so that sentences are perceptibly similar. 

While there are several approaches to text sanitisation in the literature, most do not come with code.  We adopt one that does, \textbf{CusText} from \citet{chen-etal-2023-customized}, setting $\epsilon = 10$. We add more results with other privacy budgets and discuss other approaches in the Appendix. 
Even though CusText was not designed for adversarial attacks, it works as a text sanitiser and hence we aim to study its practical consequences for labelled tasks. Outside DP, we also adopt \textbf{Presidio}\footnote{https://microsoft.github.io/presidio/} and \textbf{ZSTS} \citep{albanese2023text} sanitisers.


Finally, we also adopted \textbf{ChatGPT} as a zero-shot attacker to construct reformulations, but its performance is very weak; we discuss this in Sec~\ref{sec:ablation}.

Last, for all datasets, we used the Adam optimiser with a learning rate of 3e-5, alongside a linear schedule with a warmup ratio of 10\%. Auxiliary and target models were trained for 10 epochs for TOEFL11 and TrustPilot, and for 30 epochs for Shakespeare. The batch sizes for TOEFL11, Shakespeare and TrustPilot were 16, 8 and 64 respectively. Each experiment used a single GPU.

\subsection{Automatic Evaluation}

This is our primary evaluation, which considers how well the models being compared can produce rewritten text that behaves the same for a utility classifier while fooling a classifier that detects sensitive attributes.  

\paragraph{Attack} We evaluate the attack in terms of \textit{Attack Success} (\textbf{AS}), which, as defined in Sec \ref{sec:definition}, is built based on the outcomes for both tasks. There is a success when the adversary deceives the classifier for the privacy task, but the prediction for the utility task remains unchanged. This metric arises fairly naturally from the problem definition in Section~\ref{sec:definition}, and is similar to attack success metrics in single-task case, such as that of \citet{Jin_Jin_Zhou_Szolovits_2020}, considered the gap between the original and after-attack accuracy as the attack success, and that of \citet{garg-ramakrishnan-2020-bae}, where additionally the attack was considered a failure if all tokens were changed. We, on the other hand, need take into account the attack success under the dual-task nature of our problem.
We thus also measure the \textit{utility retention} (\textbf{UR}), which is the ratio of unchanged predictions between original and adversarial texts for the utility task. Both AS and UR are computed over original samples correctly classified.

\paragraph{Text quality} As in previous works, we analyse the quality of the adversarial texts w.r.t the original ones. We measure the cosine similarity over Universal Sentence Encoder \citep{cer-etal-2018-universal} embeddings, the ratio of the Levenshtein distance to the length of the sentences, Jaccard, Meteor and Bertscore \citep{Zhang*2020BERTScore:} metrics, as well as the proportion of tokens matching POS tags between texts. We also assess the grammar of the texts, judged by a RoBERTa-large classifier trained on the CoLA dataset \citep{warstadt-etal-2019-neural} by \citet{krishna-etal-2020-reformulating}. These measurements assess both semantic similarity and orthographic similarity, and help to evaluate how close original and adversarial samples are to each other.

\subsection{Human Evaluation}

Our secondary evaluation is in the form of two human judgement tasks regarding:

\begin{enumerate}
    \item how the sentences retain their utility by asking humans to classify the perturbed texts amongst the utility classes; and
    \item how humans rate the sentences regarding grammar and fluency.
\end{enumerate}

These first is similar to the human evaluation conducted by \citet{garg-ramakrishnan-2020-bae}, where judges had to choose the correct class label from a given set;
and the second one is modelled on that of \citet{xu-etal-2019-privacy}, where judges rate the generated sentences from 1 (``Not in the form of human language'') to 5 (``Without any grammatical error'').
We highlight that, unlike prior works, we have an explicit class to assess utility, and therefore we do not rely on abstract concepts such as ``naturalness'' \cite{garg-ramakrishnan-2020-bae}, ``meaning or natural sentence'' \cite{Jin_Jin_Zhou_Szolovits_2020} or asking humans to rank ``semantic similarity'' \cite{994d326615d645bd8d137b1b8d32540d}.

For (i), \textit{utility retention}, we randomly sampled 50 original sentences from TrustPilot and TOEFL11 datasets, alongside adversarial counterparts generated by IDT, Adv. BackTrans and TextFooler, summing to 200 sentences total. We left the Shakespeare dataset out of this study because it is unlikely that a human can effectively classify a single random sentence into a Shakespeare play, even if they have read the plays. For TrustPilot, we simply ask them to rate if the review is positive or negative. Finally, for TOEFL11, they have to discern if the sentence comes from an essay talking about \emph{cars} or \emph{students \& learning}, which are two of the possible eight topics for an essay.

We also only sampled sentences that delivered successful attacks for all algorithms. We didn't enforce the same restriction for the second task. Instead, we sampled 20 sentences from each one of the three attacks, coupled with their original counterparts, regardless of their attack success (40 * 3 = 120 sentences). We thus conducted the assignment with alongside the previous 200 sentences from task \textit{i} (320 sentences total).


We conducted both tasks on Amazon Mechanical Turk, with instructions as given in Appendix~\ref{app:humaneval}. Each sentence was rated by 3 annotators, and we paid them US\$0.13 per HIT. Each HIT takes 30 seconds, therefore they were paid around \$15.60/h, or roughly the Australian national minimum wage for the fiscal year 2023-24,\footnote{\url{https://www.fwc.gov.au/agreements-awards/minimum-wages-and-conditions/national-minimum-wage}} set at A\$23.23/h.
\section{Results}\label{sec:results}

For each dataset, we trained four models, two for privacy (one target and one auxiliary) and the other two for the utility task. Given that target and auxiliary models have their distinct training sets, which reduces the training data, we, therefore, do not aim to achieve state-of-the-art models that were trained with the whole datasets, but models strong enough to perform satisfactorily across the tasks. Table \ref{tab:target_accuracy} reports their accuracy on the same test set. We observe a very similar classification capacity for all tasks in all datasets.

\begin{table}[h]
    \centering
    \begin{tabular}{ccccc}
    \hline
         \textbf{Dataset} & \textbf{Model} & \textbf{Privacy}  & \textbf{Utility}  & \textbf{Both} \\\hline

         \multirow{2}{*}{TrustPilot\textsubscript{L}}& Target&  .5380 &  .9350 & .5080\\
         &Aux.&.5260 &.9380  & .4970 \\\cline{2-5}
         
         \multirow{2}{*}{TrustPilot\textsubscript{G}}& Target& .7298 & .9606 &.7018 \\
         &Aux.&.7228&.9586 &.6940 \\\cline{2-5}

         \multirow{2}{*}{TrustPilot\textsubscript{A}}& Target& .7862  &  .9578 & .7532\\
         &Aux.&.7932 &  .9576& .7604 \\ \cline{2-5}
         
         \multirow{2}{*}{TOEFL11} & Target & .3767 & .8405 & .3195 \\
         &Aux.& .3549 & .8391 & .3017\\\cline{2-5}
         \multirow{2}{*}{Shakespeare}& Target& .8520 & .3797 & .3280\\
         &Aux.&.8467 &.3742 & .3167\\\hline
    \end{tabular}
    \caption{Accuracy of the target and auxiliary models on the test sets. `Both' means samples were correctly classified in both tasks.}
    \label{tab:target_accuracy}
\end{table}


\subsection{Attack success}

    Table \ref{tab:attack_results} brings the results in terms of \textit{Attack Success} (\textbf{AS}) and \textit{Utility Retention} (\textbf{UR}). In summary, \attack presents competitive performance across all the datasets, showing it is a versatile technique.
    Specifically, \attack achieves the biggest UR and AS for all flavours of the TrustPilot dataset, substantially ahead of all baselines. 
    \attack is particularly effective for TrustPilot\textsubscript{L}, which holds the largest number of classes amongst the flavours.
    
    \attack stands out in the Shakespeare dataset with the highest AS and a relatively high UR. It is effective in both attacking the sentences w.r.t style and retaining their utility regarding the play they belong to. The number of classes in the utility task is the biggest amongst all tasks (17). All the other baselines, but remarkably BAE, achieve smaller AS and distort the sentences to the point of losing up to a third of their utility. The extra signal to avoid modifying important tokens for the auxiliary task appears to be important to retain the classification for the utility, especially when considering a large number of classes (17 for Shakespeare). 
    
    \begin{table}[h!]
        \centering
        \small
        \begin{tabular}{cccc}
        \hline
            \textbf{Dataset} & \textbf{Attack}  & \textbf{AS}  & \textbf{UR} \\ \hline

            \multirow{9}{*}{TrustPilot\textsubscript{L}} & IDT & \textbf{.6166}  & \textbf{1.0} \\
            &Adv. Back-translation& .1003 & .1338  \\ \cline{2-4}
            &Back-translation& .2594 &  .9782 \\
            &TextFooler&.4427 & .9872 \\
            &TextBugger&.5375 &.9875  \\
            &BAE& .4070&.9867  \\
            &Presidio& .2236 & .9868  \\
            &ZSTS& .4901 & .9133 \\
            &CusText &.4639 &.8868 \\\hline
            
            \multirow{9}{*}{TrustPilot\textsubscript{G}} & IDT & \textbf{.4544} & \textbf{.9989}\\
            &Adv. Back-translation& .1211 & .9854  \\ \cline{2-4}
            &Back-translation& .1166 & .9873  \\
            &TextFooler&.3225& .9913\\
            &TextBugger&.3113& .9932\\
            &BAE&.3215& .9923\\
            &Presidio& .0961 & .9985  \\
            &ZSTS& .3069 & .9526 \\
            &CusText &.3231 &.9406\\\hline

            \multirow{9}{*}{TrustPilot\textsubscript{A}} & IDT & \textbf{.4350}  & .9946 \\
            &Adv. Back-translation&.1563  & .9775  \\ \cline{2-4}
            &Back-translation& .1230 &  .9857 \\
            &TextFooler&.2818 & .9930 \\
            &TextBugger& .2711& .9928 \\
            &BAE& .3415&  .9872\\
            &Presidio& .0507 &  \textbf{.9959} \\
            &ZSTS& .3064 & .9397 \\
            &CusText &.2482 & .9292\\\hline

            \multirow{9}{*}{TOEFL11} & IDT & .4485 & .9716\\
            &Adv. Back-translation& .5183 &  \textbf{.9762} \\ \cline{2-4}
            &Back-translation&.4445  &  .9652 \\
            &TextFooler&.4293& .9525\\
            &TextBugger&.4346&.9694 \\
            &BAE&.4807& .9542\\
            &Presidio& .3843 & .9552  \\
            &ZSTS& .4441 & .6678 \\
            &CusText &\textbf{.5739} & .7466\\\hline
            \multirow{9}{*}{Shakespeare} & IDT & \textbf{.6812} &\textbf{.8531} \\
            &Adv. Back-translation&  .2757 &  .8492  \\ \cline{2-4}
            &Back-translation&.2575  &  .6654 \\
            &TextFooler&.4626&.7149 \\
            &TextBugger&.4774& .7511\\
            &BAE&.4256& .6551\\
            &Presidio& 0.0 & .4927  \\
            &ZSTS& .0754 & .3064 \\
            &CusText &.1218 & .3275\\\hline
        \end{tabular}
        \caption{\textit{Attack Success} (\textbf{AS}) and \textit{Utility Retention} (\textbf{UR}) for each dataset under different attacks.}
        \label{tab:attack_results}
    \end{table}

    For TOEFL11, \attack has the fourth largest AS, behind only BAE, Adv. Back-Translation and CusText, and the second biggest UR. We note here the potential advantages of back-translation: without the signal from the adversary, the AS is (a quite decent) 44.45\%, but the signal improves it to 51.83\%. However, keeping the stability of the translator is hard, and it is the worst performer in TrustPilot (location), mostly due to mode collapse. The extra signal from the adversary makes the training stability an extra challenge.

    In terms of the principal comparator, Adv. Back-Translation, the reasons for its relatively poorer performance vary.  For TrustPilot\textsubscript{L}, it has trouble retaining accurate predictions on the utility class, which is what brings down its attack success there.  However, for the others, it has a high UR; the difficulty is in changing the texts sufficiently to obscure the sensitive attribute.

    The sanitiser algorithms, namely Presidio and ZSTS, are amongst the worst performers in general. Presidio aims to modify PII tokens which may have little effect in classification tasks, whereas ZSTS has the opposite effect, by modifying rare words and thus also harming utility.

    \attack also outperforms the TextAttack algorithms (TextFooler, TextBugger and BAE) across all datasets. These baselines show more stable behaviour compared to back-translation and sanitisers, but they still miss the utility task explicit signal, which makes a difference for \attack.

    Details on number of successful attacks are in Appendix~\ref{app:num-adv-texts}.

\subsection{Quality of adversarial texts}

    We show adversarial examples generated by the attacks on Table \ref{tab:examples_texts}, alongside their labels.    
   
    \begin{table}[t]
        \centering
        \footnotesize
        \begin{tabular}{cc}
        \hline
         \textbf{Attack} & \textbf{Text} \\
        \hline
        --&I would certainly recommend parcel hero and use their services again\\
             \attack&I would certainly recommend parcel \textbf{entertainer} and use their services again \\
             TF&I would certainly \textbf{commended} parcel hero and use their services again \\
             Adv. Back-trans &\textbf{i} would \textbf{definitely} recommend parcel hero and use their services again  \\
             Back-trans. &I would certainly recommend \textbf{a package} hero and use their services again.\\
             TB&I would certainly \textbf{reCommend} parcel hero and use their services again\\
             BAE&I would certainly \textbf{marry} parcel hero and use their services again\\
             CusText & I would \textbf{yet} recommend parcel \textbf{fame} and use their services \textbf{leaving} \\ 
             Presidio &I would certainly recommend parcel hero and use their services again\\
             ZSTS &i would \textbf{simply join some deliveries not} use \textbf{these weapons} - \\
             \hline

              -- &Of course , the fact will be useless when the main idea is not transported to the students .  \\
             \attack & Of course , the fact will be useless when the main idea is not \textbf{traced} to the students .  \\
             Adv. Back-trans & \textbf{of} course, the fact will be useless when the main idea is not transported to students.  \\
             Back-trans. &Of course, the fact will be useless when the main idea is not carried to the students.\\
             TF & \textbf{With} course , the fact will be useless when the main idea is not transported to the students . \\
             TB & \textbf{Del} course , the fact will be useless when the main idea is not transported to the students .  \\
             BAE &\textbf{of} course , the fact will be useless when the main idea is not transported to the students .   \\
             CusText & Of course \textbf{with} the \textbf{what} will be useless \textbf{again} the main idea is \textbf{if} transported \textbf{its} the students \textbf{for} \\
             Presidio &Of course , the fact will be useless when the main idea is not transported to the students .\\
             ZSTS &\textbf{of} course , \textbf{any solution} will be \textbf{reached so }the \textbf{good system }is not \textbf{visible} to the \textbf{lab} .\\
             \hline

             -- & no more than my staying here in rome might mean to you in egypt. \\
             \attack & no more than my staying here in rome might mean \textbf{up} you in egypt.  \\
             Adv. Back-trans. & no more than my \textbf{stay} here in rome might mean to you in egypt. \\
             Back-trans. &No more than my \textbf{stay} here in Rome could mean for you in Egypt.\\
             TF & no more than my staying here in rome might \textbf{intentioned} to you in egypt. \\
             TB & no more than my staying here in rome might \textbf{meaning} to you in egypt. \\
             BAE & no more than my staying here in rome might \textbf{render} to you in egypt.  \\ 
             CusText & \textbf{nor} more \textbf{less} \textbf{something} staying \textbf{close} \textbf{on} rome might \textbf{exactly} to you in egypt.\\
             Presidio &no more than my staying here in rome might mean to you in egypt.\\
             ZSTS &no more \textbf{everything you living }here in \textbf{greece must} mean \textbf{having study }in \textbf{fact} .\\
             \hline
             
        \end{tabular}
        \caption{Examples of adversarial sentences generated by the attacks. The top samples come from TrustPilot dataset; the middle ones are from TOEFL11 and the bottom one are Shakespeare sentences.}
        \label{tab:examples_texts}
    \end{table}

    \begin{table}[h!]
        \setlength{\tabcolsep}{1pt}
        \centering
        \scriptsize
        \begin{tabular}{cccccccccccc}
        \hline
            \textbf{Dataset} & \textbf{Attack}  & \textbf{\makecell{Matching \\ POS tags}}  & \textbf{\makecell{Grammar \\ correct}} & \textbf{\makecell{Cosine \\sim.}} & \textbf{\makecell{Levenshtein \\ ratio}} &\textbf{\makecell{Changed \\ words}} & \textbf{\makecell{Jaccard \\ sim.}}& \textbf{Meteor}& \textbf{\makecell{BertScore \\ F1}}  &\textbf{BLEU}& \textbf{RougeL}  \\\hline

            \multirow{9}{*}{TrustPilot\textsubscript{L}} & IDT & \textbf{1.0} & .7222 &.2341 & .9614& \textbf{.0570}& \textbf{.8945}  &.9440&.9756 &.9189 &.9439\\
            &Adv. Back-trans. &0.0 &.0019&.0219 &.1803 &10.42&.0049 & .0135& .7133 & 0.0&.0034\\
            &Back-trans. &.0237&.7782&.2276&.8591&.7056&.6039&.8143&.9600 &.4779 &.7952\\
            &TextFooler& .5776 & .6794 & .2289 & .9402 &.0823 & .8466  &.9232&  .9757 &.8716 &.9193\\
            &TextBugger& .6575 & \textbf{.8200} & .2372  &  \textbf{.9827}& .0989 &  .8896 &\textbf{.9483}&.9742 & .9066&.9388\\
            &BAE& .6460 & .7566 & .2319 & .9490  & .0706 & .8632  &.9332& \textbf{.9804} &.8887 &.9292\\
            &CusText &.0030&.0251&.2023&.7800&.5147 &.4866&.6079&.9092  & .3654&.6301\\
            &Presidio &0.0&.8157&\textbf{.2479}&.9567&\textbf{.0331}&\textbf{.9387}&\textbf{.9615}&.9745 &\textbf{.9208} &\textbf{.9547}\\
            &ZSTS &.0078&.1968&.1798&.6507&.7486&.2943&.5255&.8789 &.3285 &.5413\\
            \hline
            
            \multirow{9}{*}{TrustPilot\textsubscript{G}} & IDT & \textbf{1.0} &.7039 & .2516&.9592&\textbf{.0549}&\textbf{.8952} &\textbf{.9450}& .9795 &\textbf{.9135} &.9447\\
            &Adv. Back-trans. & .0351& .8694& .2515& .8644&.6611&.6121 &.8307 &.9665  &.5122 &.8317\\
            &Back-trans. &.0343&.7974&.2502&.8499&.7033&.6256&.8019&.9624 &.5409 &.7835\\
            &TextFooler&.5854 &.6778 & .2532 & .9395 &.0843 &.8424  &.0702&.0195  &.8745 &.9168\\
            &TextBugger& .4351& \textbf{.7650}& \textbf{.2615}  &\textbf{ .9689} & .1470& .8446 &.9262&.9692 & .8716&.9145\\
            &BAE& .5850&.7580 &.2555 & .9483 &.0766 &.8556  &.9293& \textbf{.9811} &.8721 &.9269\\
            &CusText &.0109&.0350&.2322&.7982&.4836&.5194&.6470&.9116&.4001&.6627 \\
            &Presidio &.0014&\textbf{.8809}&.2593&.9569&.1107&\textbf{.9309}&.9598&.9768 & .9110&\textbf{.9595}\\
            &ZSTS &.0017&.1783&.1930&.6341&.7646&.2613&.5241&.8755 &.2442 &.5408\\
            \hline

            \multirow{9}{*}{TrustPilot\textsubscript{A}} & IDT & \textbf{1.0} & .6787 &.1914 &.9378 & \textbf{.0930}&\textbf{.8461}&.8999&.9726 &\textbf{.9158} &.9098 \\
            &Adv. Back-trans. &.0514 &.8776 &.2225 &.8473 &.6671&.5983 &.7994 & .9647 &.4586 &.8148\\
            &Back-trans. &.0438&.8043&.2227&.8377&.7932&.6077&.7809&.9600 &.4912 &.7680\\
            &TextFooler& .5234 & .6308 & .2233 &.9180  &.1111 &.8009  &.8934&.9698  &.8609 &.8907\\
            &TextBugger&.3411  & .7357 &  \textbf{.2388} & \textbf{.9583} &.2092  &.8089  &\textbf{.9048}& .9635 &.8528 &.8930\\
            &BAE& .5439 &\textbf{.7396 } &.2208  & .9284 & .1014 & .8214 &.8982&\textbf{.9754}  & .8754& .9012\\
            &CusText &.0227&.0417&.2076&.7949&.5137&.5180&.6403&.0268  & .3995&.6602\\
            &Presidio &.0010&\textbf{.8864}&\textbf{.2565}&.9530&.1261&\textbf{.9295}&\textbf{.9575}&\textbf{.9760} & .9143&\textbf{.9554}\\
            &ZSTS &.0063&.2049&.1750&.6304&.7907&.2578&.5147&.8749 &.2576 &.5338\\
            \hline

            \multirow{9}{*}{TOEFL11} & IDT & \textbf{1.0} &.4756&.1028 &.9467 &.0755&\textbf{.8739} &.9310&.9731 &.8532 &.9257\\
            &Adv. Back-trans. & .0755&.8652 &.1055 &.8835&.6396& .5013& .8404&.9585  & .5554&.8130\\
            &Back-trans. &.0600&.8452&.1043&.8514&.6711&.4882&.7916&.9509 & .4961&.7511\\
            &TextFooler&.7839 &.4565 & .1053& .9438&.0801 &.8582  &.9349&.9792  &.8582 &.9201\\
            &TextBugger&.5237 &.5535 &\textbf{.1066}& \textbf{.9758 }&.1026 &.8717  &.9427&.9735 &.8756 &.9218\\
            &BAE&  .8565&\textbf{.5898} &.1049 &.9557 &\textbf{.0722}  & .8733  &\textbf{.9473}&\textbf{.9860} & \textbf{.8778}&.9373\\
            &CusText &.0021&.0081&.0877&.7340&.5934&.4298&.5434& .8868 & .2327&.5651\\
            &Presidio &.0037&\textbf{.6455}&.0899&.9184&.1609&\textbf{.8871}&.9276&.9623 & .8113&\textbf{.9332}\\
            &ZSTS &.0443&.3513&.0924&.6781&.5670&.3959&.5734&.9031 &.2934&.5775\\
            \hline
            
            
            \multirow{9}{*}{Shakespeare} & IDT & \textbf{1.0} &.4188&.0948 & .8822&.1872&.7015 &.7846&.9505 &.7579 &.8160 \\
            &Adv. Back-trans. & .1130&.7470 &\textbf{.0968} & .8928&.5535& .6514&.7961 & .9642 &.6804 &.8136\\
            &Back-trans. &.1087&.7484&.0911&.7727&.8460&.3816&.7001&.9402 & .3710&.6773\\
            &TextFooler&.5656 &.3744 & .0888  &.8534 &.2069 & .6750 &.7769&.9496  &.7032 &.7992\\
            &TextBugger&.5505 & \textbf{.6221}& .0964 & \textbf{.9422}  &\textbf{.1747}&\textbf{.7530}  &\textbf{.8599}&\textbf{.9606} &.7869 &.8524\\
            &BAE& .5505& .5589 &.0880 &  .8263 &.2408 &.6343 &.7241&.9427 &.6879 & .7615\\
            &CusText &.0689 &.0923 &.0827  &.7789&.4094 &.4911 &.5807&.9210  &.3832 &.6479\\
            &Presidio &0.0&\textbf{.7536}&\textbf{.1365}&.8649&\textbf{.0993}&\textbf{.8239}&\textbf{.9103}&.9516 &\textbf{.7925} &\textbf{.9068}\\
            &ZSTS &.0975&.3963&.0832&.6046&.7512&.2194&.4782&.8762 & .2294&.4724\\
            \hline
        \end{tabular}
        \caption{Quality measurements of adversarial texts against the original ones.}
        \label{tab:sentence_quality}
    \end{table}

    For TrustPilot, we note that all attacks chose the same word to be replaced, again with the exception of CusText. For Shakespeare, all TextAttack algorithms picked the same word to attack.
    
    For the TOEFL11, the original sentence, written by a Chinese learner, was modified in ways to deceive the target model to classify it as written either by a French or an Arabic speaker. The topic, ``if it is more important for students to understand ideas and concepts than it is for them to learn facts'', shortened in the table as ``Students \& learning'', was kept. In addition, all attacks chose the same word to change for the Shakespeare sentences except \attack, which found a different one, and CusText, which heavily modified the sentence.

    Table \ref{tab:sentence_quality} shows different quality measurements for the generated sentences. 
    The key takeaway is that the text quality of IDT is generally good --- that is, the rewritten sentences were close to the original --- and comparable to the (single-task) adversarial attacks.
    In general, adversarial texts generated by TextBugger present the best results. However, this algorithm had the poorest performance amongst all attacks, which is an indication that its changes were too subtle. This is corroborated by our grammar checker which deems TextBugger texts as the most correct in general --- and TextFooler produces the smallest number of correct texts for all datasets. \attack holds the highest Jaccard similarity (which also correlates well with the percentage of perturbed words) between adversarial and original texts in four out of the five datasets, on top of also being the attack with the highest AS in four datasets. 

    We notice that the adversarial training of \citet{xu-etal-2019-privacy} is heavily inspired on Generative Adversarial Networks \citep{NIPS2014_5ca3e9b1}, which commonly face instability training issues. While training our Adv. Back-Translator, we have encountered mode collapse, as well as memory issues while training, since two neural networks must be held in memory. We found particularly challenging finding a balance in some cases, as in TrustPilot (location), for which mode collapse produced the poorest quality metrics. 

    We also observe how far the reformulations of CusText, as an example of text sanitisation, are from the other algorithms. Despite using a reasonable large $\epsilon=10$, very few adversarial sentences match their POS tags with their original counterparts, possibly a consequence of the large percentage of perturbed words. CusText is also the the worst performer in terms of AS success, mainly driven by its lower utility retention.

    We also notice the importance of the constraint of ensuring the same POS tags between original and adversarial tokens. The baselines, which do not enforce this strict constraint, generate around half of the texts with different part-of-speech. The cosine similarity, a commonly used constraint by the baselines, is not enough to prevent mismatching POS tags.


\subsection{Ablation and further studies}
\label{sec:ablation}
    
    \paragraph{Different target model} We study how transferable \attack and TextAttack baselines are when the auxiliary models have a distinct architecture from the target model. The target is a distilled GPT2, whilst the auxiliaries are distilled RoBERTa. We also report the differences between the \textbf{AS} and the \textbf{UR} reported in Table \ref{tab:attack_results} in Table \ref{tab:attack_results_differ_architerctures}.

    \attack is the clear winner when transferability is taken into account, and the baselines are much more reliant on the architecture to find words to substitute. \attack achieves the highest AS in all datasets but TrustPilot\textsubscript{G}, which is ranks second by a tiny margin of 0.15\% behind BAE.

    \begin{table}[h]
        \centering
        \footnotesize
        \begin{tabular}{cccc|cc}
        \hline
            \textbf{Dataset} & \textbf{Attack}  & \textbf{AS} & \textbf{Diff.}  & \textbf{UR} & \textbf{Diff.} \\ \hline

            \multirow{4}{*}{TrustPilot\textsubscript{L}}  & IDT &  \textbf{.3947} & -.2219 & \textbf{ 1.0} & .0000 \\
            &TextFooler& .3608& -.0819 & .9877 & .0005 \\
            &TextBugger& .3663& -0.1712 & .9969 &.0094  \\
            &BAE& .3023& -0.1047 & .9941 & .0074 \\\hline
            
            \multirow{4}{*}{TrustPilot\textsubscript{G}}  & IDT & .2623 & -.1921& \textbf{1.0} & .0011\\
            &TextFooler&.2415 & -.0810&.9927 &.0014 \\
            &TextBugger&.2076 &-.1037 & .9941&.0009 \\
            &BAE&\textbf{.2638} &-.0577 &.9915 &-.0008 \\\hline

            \multirow{4}{*}{TrustPilot\textsubscript{A}}  & IDT &  \textbf{.3758} & -.0592 & \textbf{.9957}  & .0011 \\
            &TextFooler&.2844  & .0026 &  .9930 & .0000 \\
            &TextBugger& .2737 & .0026  &.9928   &.0000  \\
            &BAE& .3510 & .0095 & .9892 & .0020 \\\hline

            \multirow{4}{*}{TOEFL11} & IDT & \textbf{.4210} &-.0275& \textbf{.9845} &.0129\\
            &TextFooler&.3589 &-.0704 & .9549& .0024\\
            &TextBugger&.3702 &-.0644 & .9691& -.0003\\
            &BAE& .3843&-.0964 &.9559 &.0017 \\\hline
            
            
            \multirow{4}{*}{Shakespeare} & IDT & \textbf{.4022} & -.2790 &\textbf{.7873} &   -.0658 \\
            &TextFooler&.3519 &-.1107 & .7111&-.0038 \\
            &TextBugger&.3245 &-.1529 &.7327 &-.0184 \\
            &BAE&.3539 & -.0717&.6234 &-.0317\\\hline
        \end{tabular}
        \caption{\textbf{AS} and \textbf{UR} for target and auxiliary models with \textbf{different architectures}, alongside their differences to the setting reported in Table \ref{tab:attack_results}.}
        \label{tab:attack_results_differ_architerctures}
    \end{table}

    \paragraph{Relaxing constraints} We examined relaxing the constraint of forcing adversarial texts to have the same POS tags for every token in the original text. More specifically, we re-design \attack to allow nouns to be swapped by verbs and vice-versa, as in the TextAttack implementation of TextFooler and BAE. As in TextAttack, we also use the `universal' tagset of NLTK library \citep{bird-loper-2004-nltk} to compute the POS tags. Table \ref{tab:attack_results_constraints} shows the results.

    Except for TOEFL11, relaxing the constraints helps the AS for all datasets. For Shakespeare, it is particularly helpful to make the AS above 70\%, the highest results across all our experiments.

    \paragraph{Ignoring important tokens for utility} We modify \attack to ignore all words for the utility task, to make the attack behaviour akin to the attacks from the literature. Then, we measure the effects of utility retention and the predictions for the privacy tasks. Results are shown in Table \ref{tab:attack_results_constraints}.

    We see that such relaxation yields a negative impact for all datasets in terms of AS. The sharpest decrease happens for Shakespeare sentences.

    \begin{table}[h]
        \centering
        \small
        \begin{tabular}{cccc|cc}
        \hline
            \textbf{Dataset} & \textbf{Attack}  & \textbf{AS} & \textbf{Diff.}  & \textbf{UR} & \textbf{Diff.} \\ \hline
            
            \multirow{2}{*}{TrustPilot\textsubscript{L}}  & IDT\textsubscript{s} &  .5966  &-.0200   & 1.0  &.0000 \\
            &IDT\textsubscript{u}& .6067 &-.0099  &1.0&.0000   \\\hline
            
            \multirow{2}{*}{TrustPilot\textsubscript{G}}  & IDT\textsubscript{s} &.4409   &.0135   & .9978  &-.0011\\
            &IDT\textsubscript{u}& .4305 & -.0239 & .9957 &  -.0032 \\\hline

            \multirow{2}{*}{TrustPilot\textsubscript{A}}  & IDT\textsubscript{s} &  .4650  & .0300  & .9986  &.0040 \\
            &IDT\textsubscript{u}& .4304 &-.0046& .9933& -.0013 \\\hline

            \multirow{2}{*}{TOEFL11} & IDT\textsubscript{s} & .4293  & -.0192  & .9662  &-.0054\\
            &IDT\textsubscript{u}& .4342 &-.0143& .9578 & .0036  \\\hline
            
            
            \multirow{2}{*}{Shakespeare} & IDT\textsubscript{s} &   .7177&  .0365 &  .8742 &.0211\\
            &IDT\textsubscript{u}&.5476 &-.1336 &.7574 &-.0957 \\\hline
            
        \end{tabular}
        \caption{\textbf{AS} and \textbf{UR} for \attack under different constraints, alongside their differences to the setting reported in Table \ref{tab:attack_results}. \textbf{IDT\textsubscript{s}} means verbs can be swapped by nouns and vice-versa, and \textbf{IDT\textsubscript{u}} ignores the utility task to sample replacement words.}
        \label{tab:attack_results_constraints}
    \end{table}

\begin{table}[h]
        \centering
        \small
        \begin{tabular}{cc|cc}
        \hline
            \textbf{Measure} & \textbf{Value} & \textbf{Measure} & \textbf{Value} \\
            \hline
            Matching POS tags  & .0208 & Grammar correct & .8750\\
            Cosine sim. & .1219 & Levenshtein ratio &.5569 \\
            \% Perturbed words & .9330 & Jaccard sim. & .1405 \\ 
            Meteor & .3201 & BERTScore F1 & .9065 \\
            BLEU & .0602& RougeL & .3964 \\\hline
        \end{tabular}
        \caption{Measurements on the adversarial sentences generated by ChatGPT.}
        \label{tab:chatgpt}
    \end{table}

    \paragraph{ChatGPT reformulations} We selected the intersection of successfully attacked Shakespeare sentences by all attacks and asked ChatGPT to rewrite them. Specifically, our prompt is:

    ``\textit{Given the following text, rewrite it in a way that makes it different from <old/modern> style, but keep it in the style of a <play name> play.}''

    We then measured the effect of the reformulations on the predictions by the target models.
    Individually, regarding the privacy attribute style, ChatGPT deceived the victim 34 times. For the utility, play, the predictions are preserved also 34 times, but not for the same sentences. Overall,  ChatGPT deceives the classification w.r.t. the style while keeping the predictions for the play in 24 out of 48 instances. 
    
    Considering that these are sentences that all algorithms successfully attacked, ChatGPT's performance of 50\% can be deemed low, which reinforces the need for fine tuning if using the model for more complex tasks like dual task attacks.

    Qualitatively, its adversarial sentences are poorer than all other attacks according to the measurements from Table \ref{tab:chatgpt}. ChatGPT is a generative model, and signals to enforce constraints on its output have to be fed in the prompt, which is not always reliable.
    In addition, ChatGPT is verbose and may produce texts longer than the inputs, which makes the percentage of perturbed words above 100\%, and matching POS tags near zero. These become clear to understand from Table \ref{tab:chatgpt_samples}, where we show a few sentences rewritten by the model.

       \begin{table}
        \footnotesize
        \begin{tabular}{p{0.47\linewidth} | p{0.47\linewidth}}
        \hline
            \textbf{Original} & \textbf{Adversarial} \\
            \hline
            god bless you, bottom, god bless you. & Hark, may divine favor grace thee, Bottom, may divine favor grace thee!\\\hline
            no more than my staying here in rome might mean to you in egypt. & No more than my sojourn in Rome might signify to thee in Egypt.\\\hline
            what do you mean, do i think so? & What signify you? Do I hold such a belief?\\\hline
            her name is portia. &Her appellation doth resonate as Portia. \\\hline
            do you intend to walk outside? & Dost thou purpose to traverse the open precincts?\\ \hline
        \end{tabular}
        \caption{Examples of ChatGPT reformulations.}
        \label{tab:chatgpt_samples}
    \end{table}
    
\subsection{Human evaluation results}

We report the human evaluation regarding utility retention and grammaticality and fluency judgements.  In this, we just compare our model \attack with the principal baseline, adversarial backtranslation, and the TextFooler adversarial attack that is at the base of \attack.

\paragraph{Utility evaluation}

We asked annotators to classify the sentences with respect the utility task. For TOEFL11, it means judging whether the sentence talks about ``cars'' or ``students \& learning'', and for TrustPilot it means rating it as a ``good'' or ``bad'' review. We report the accuracy of the annotators in Table \ref{tab:utility_human_eval}.

\begin{table}[h!]
    \centering
    \begin{tabular}{ccccc}
    \hline
        \multirow{2}{*}{\textbf{Dataset}} & \multicolumn{4}{c}{\textbf{Utility accuracy (\%)}}\\
         & Original & TextFooler & Adv. BackTrans. & IDT\\ \hline
        TOEFL11 & 93.33 & 94.66 & 94.66 &95.33 \\
        TrustPilot & 100.0  & 96.90 & 92.13 & 96.11\\ \hline
    \end{tabular}
    \caption{Human evaluation results for topic (TOEFL11) and sentiment (TrustPilot) classification.}
    \label{tab:utility_human_eval}
\end{table}

In general, humans can discern with ease the class of the sentences for all algorithms. Numerically, though, IDT achieves the highest utility accuracy score for the TOEFL11 dataset, with a score of 95.33\%. This surpasses even the original sentences (93.33\%) and other transformation methods like TextFooler and Adv. BackTrans. (both at 94.66\%).




\paragraph{Grammar and fluency} The second assignment asked annotators to judge the sentences with respect to their grammar and fluency. Their judgements are summarised in Table \ref{tab:grammar_human_eval}.

\begin{table}[h!]
    \centering
    \begin{tabular}{ccccc}
    \hline
        \multirow{2}{*}{\textbf{Dataset}} & \multicolumn{4}{c}{\textbf{Grammar and fluency (1-5)}}\\
         & Original & TextFooler & Adv. BackTrans. & IDT\\ \hline
        TOEFL11 &  3.91 & 3.59 & 3.98 & 3.36 \\
        TrustPilot & 4.16 & 3.84 & 4.26 & 3.80\\ \hline
    \end{tabular}
    \caption{Human evaluation results for grammar and fluency. The closer to 5, the better quality the sentences hold according to the annotators.}
    \label{tab:grammar_human_eval}
\end{table}

For all algorithms, as well as the original sentences, across both datasets, scores are mostly between 3 and 4 in our 1-5 Likert scale, with a couple above 4 (higher scores being better).  A score of 3 indicates basically fluent and understandable, and a score of 4 that there are only one or two minor errors.

We notice that all algorithms achieve better grades in TrustPilot than TOEFL11. This is explained by the fact that TOEFL11 sentences are written by learners of English and taken from long essays. Therefore are more prone to contains errors than TrustPilot reviews, although the original sentences here also score quite below 5 (free of grammatical errors) as these are often quickly written product reviews that have not been proofread. 
In terms of methods, the adversarial backtranslator clearly produces the most fluent texts.
Interestingly, in fact, the adversarial backtranslator achieves better scores than the original sentences in both datasets perhaps because, as a generative model, it is fixing spelling mistakes by learners. This doesn't happen with the other methods, which are token-based replacements.  And although outperformed on this property by the adversarial backtranslator, the IDT scores still indicate an acceptable degree of fluency.


\section{Conclusions}\label{sec:conclusions}

Inference of the private attributes of writers, such as age or native language, are detectable with increasing ease by machine learning models in scenarios where text is made public, such as product reviews or social media forums.
In this paper, we have proposed a method based on adversarial attacks for rewriting text to prevent detection of such private attributes, while maintaining the utility with respect to some primary task (e.g. sentiment classification).

We have shown that most existing privacy-protecting methods that rewrite text, such as text sanitisation, do not already solve this specific problem.  We have further shown that \attack performs the best against several baselines, including generative models with the same goal, in terms of ability to preserve privacy while maintaining utility; and it manages to keep texts fairly similar to the original while doing so.  This makes it a strong baseline for future work on a task that we consider quite important.

There are several directions for future work.  
\attack, or approaches similar to it, could be improved by starting from a different adversarial attack, using different constraints, applying different methods for identifying words to be changed, and so on.  One particular aspect that could be a focus of improvement is in human-judged fluency, where the generative baseline, adversarial backtranslation, was clearly superior; while this was not so important in the kinds of scenarios considered in our work, it would be in scenarios where human inspection of the text is a factor.  Generative methods that can avoid problems like mode collapse are thus also an important future direction.




\appendix\label{appendix}

\appendixsection{Annotation Guidelines}\label{app:humaneval}

\paragraph{Utility retention}

 What is this sentence talking about? 

 <<Sentence>>

 \begin{itemize}
     \item Students \& Learning.
     \item Cars.
 \end{itemize}

\paragraph{Grammar and fluency}  Please judge the text according to its grammar and fluency.

\begin{enumerate}
    \item Not in the form of human language.
    \item Can not understand what is the meaning, but is still in the form of human language.
    \item Basically fluent and has three or more minor grammatical errors or one serious grammatical error that does not have strong impact on understanding.
    \item Fluent and has one or two minor grammatical error that does not affect understanding.
    \item Without any grammatical error.
\end{enumerate}

\appendixsection{Data splits}

    We split each dataset into target and auxiliary model data as shown in Table \ref{tab:dataset_splits}. TOEFL11\textsubscript{S} refers to the TOEFL11 experiments described in the main body of this paper, and TOEFL11\textsubscript{C} is about the extra experiments from Appendix \ref{app:toefl}

    \setlength{\tabcolsep}{3pt}
    \begin{table}[h!]
        \centering
        \footnotesize
        \begin{tabular}{cccc|cc}
        \hline
        \textbf{Dataset} &
        \textbf{Train} &
        \textbf{Valid.} &
        \textbf{Test} & \textbf{Train} &
        \textbf{Valid.}  \\ \hline
         
         TrustPilot\textsubscript{L} & 5,960 & 663 & 1,000 & 5,959& 663 \\
         TrustPilot\textsubscript{G} & 100,000 & 1,000 & 5,000 & 100,000& 1,000 \\
         TrustPilot\textsubscript{A} & 85,745 & 1,000 & 5,000 & 85,745& 1,000 \\

         TOEFL11\textsubscript{S} & 72,301 & 16,146 & 15,951 & 65,070  & 7,231 \\
         TOEFL11\textsubscript{c} & 14,214 & 3,182 & 3,105 &  12,793 & 1,422\\
         Shakespeare & 23,491  & 2,611 & 4,000 & 23,491  & 2,611 \\
         \hline
        \end{tabular}
        \caption{\textmd{Data splits for target (left) and auxiliary models (right). TrustPilot\textsubscript{\textbf{L}, \textbf{G}, \textbf{A}} refer to \textbf{L}ocation, \textbf{G}ender and \textbf{A}ge views of this dataset.}}
        \label{tab:dataset_splits}
    \end{table}

\appendixsection{Implementing the Adversarial Back-Translation baseline}\label{app:backtranslation}

We reimplemented the back-translation with adversarial training proposed by \citet{xu-etal-2019-privacy} in 2019, with a few changes since more powerful architectures have been developed since then. 

Their framework involves three steps: (i) using two translators (English to French, followed by French to English) to turn the datasets labelled with sensitive attributes into parallel corpus; (ii) train a classifier and a back-translator in an adversarial fashion using the parallel corpus from the step before; (iii) use the back-translator to generate adversarial sentences.

In their study, \citet{xu-etal-2019-privacy} trained a translator for step (i) using the Europarl v7 dataset from \citet{koehn-2005-europarl}. The authors reported a BLEU score of 36.24\%. We used the English to French and French to English models provided by \citet{TiedemannThottingal:EAMT2020} to translate the texts from our datasets. As explained in Section \ref{par:baselines}, we used two MarianMT pretrained models \citep{TiedemannThottingal:EAMT2020} to generate the parallel corpus.

For step (ii), the authors trained the Transformer from \citet{10.5555/3295222.3295349}. We instead fine tuned a pretrained BART \citep{lewis-etal-2020-bart}. For the first 10 epochs we only trained the adversary classifier (a BART classification head). Then, for the next 10 epochs, we only fine-tuned the translator. Finally, we trained both for 14 epochs. Our learning rate was set to 1e-5, batch size of 32 (except for Toefl, for which use used 16) and Adam optimiser for both translator and adversary. \citet{xu-etal-2019-privacy} reported they used the mean of the hidden representations of the Transformer as input to the classifier. We found more useful to take the last hidden state of BART.

\appendixsection{Experiments with longer texts}\label{app:toefl}

Since the TOEFL11 dataset contains longer essays than the maximum input size of base Transformer models (usually 512 or 1024), we organised an extra set of experiments for which we split the texts into chunks of 5 sentences. Hereafter, we call this view of the TOEFL11 as TOEFL11\textsubscript{C}, while the other one, described in the main body of the paper, we refer now as TOEFL11\textsubscript{S}. Accuracy of the target and auxiliary models on the test sets are depicted in Table \ref{tab:toeflc_testset}.

\attack appears to have more difficulty in dealing with the longer texts from TOEFL11\textsubscript{C}, made of chunks of sentences, than the single sentences of TOEFL11\textsubscript{S}. This phenomenon is also observed for all baselines.

\begin{table}[h]
    \centering
    \begin{tabular}{ccccc}
    \hline
          \textbf{Dataset} &\textbf{Model} & \textbf{Privacy}  & \textbf{Utility}  & \textbf{Both} \\\hline
         \multirow{2}{*}{TOEFL11\textsubscript{C}}  &Target& .5574 & .9742 & .5443\\
         &Aux.& .5539&.9784 & .5407\\\hline         
    \end{tabular}
    \caption{Accuracy of the target and auxiliary models on the test sets of TOEFL11\textsubscript{c}. `Both' means samples were correctly classified in both tasks. }
    \label{tab:toeflc_testset}
\end{table}

            

\begin{figure}[h]
    \centering
    \small
    \begin{tikzpicture}
        \begin{axis}[
            ybar,
            bar width=0.4cm,
            width=0.5\textwidth,
            height=6cm,
            ylabel={\%},
            symbolic x coords={IDT, TextFooler, TextBugger, BAE},
            xtick=data,
        ]
            \addplot[fill=blue!30] coordinates {(IDT, 29.60) (TextFooler, 32.25) (TextBugger, 31.13) (BAE, 27.27)};
            \addlegendentry{Same architectures}
            
            \addplot[fill=red!30] coordinates {(IDT, 34.16) (TextFooler, 25.29) (TextBugger, 27.05) (BAE, 28.51)};
            \addlegendentry{Differing architectures}
        \end{axis}
    \end{tikzpicture}
    \caption{Attack Success rates against TOEFL11\textsubscript{c}. }\label{fig:toeflc_attack}
\end{figure}
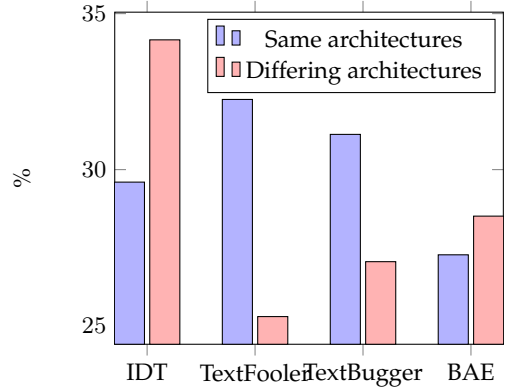

Figure \ref{fig:toeflc_attack} shows the Attack Success rates when target and auxiliary models have the same (blue) and distinct (pink) architectures. Under the first setting, \attack is outperformed by TextFooler and TextBugger, but it stays ahead of BAE. However, \attack achieves the biggest AS when architectures differ, showing stronger transferability. 

Regarding the quality of the adversarial texts, Table \ref{tab:sentence_quality_toeflc} shows that \attack performs better in qualitative terms: it keeps the largest Jaccard similarity, and now it also has the least percentage of perturbed words and biggest Meteor, surpassing BAE from Table \ref{tab:sentence_quality}.

   \begin{table}[h!]
        \setlength{\tabcolsep}{1pt}
        \centering
        \footnotesize
        \begin{tabular}{cccccccccc}
        \hline
            \textbf{Dataset} & \textbf{Attack}  & \textbf{\makecell{Matching \\ POS tags}}  & \textbf{\makecell{Grammar \\ correct}} & \textbf{\makecell{Cosine \\sim.}} & \textbf{\makecell{Levenshtein \\ ratio}} &\textbf{\makecell{\% Perturbed \\ words}} & \textbf{\makecell{Jaccard \\ sim.}}& \textbf{Meteor}& \textbf{\makecell{BERTScore \\ F1}}  \\ \hline
            
            \multirow{4}{*}{TOEFL11\textsubscript{C}} & IDT & \textbf{1.0} &.4587 & .2562&.9808 & .0261&.9477 &.9741 &.9857 \\
            &TextFooler& .5849& .4174&.2595 &.9724 &.0348 & .9145&.9688 &.9864 \\ 
            &TextBugger& .3977& .4981&.2598 &.9891 &.0988 &.9249 &.9733 & .9821\\
            &BAE& .6320 &.4528 & .2614&.9745 &.0375 &.9137&.9696&.9882\\\hline
            
        \end{tabular}
        \caption{Measurements on the quality of adversarial texts against the original ones from TOEFL11\textsubscript{C}.}
        \label{tab:sentence_quality_toeflc}
    \end{table}

\appendixsection{Further CusText evaluation}\label{app:dp_experiments}

   We conducted an experiment with CusText algorithm to create a reformulated text based on differential privacy noise. The method, proposed by \citet{chen-etal-2023-customized}, is a text sanitisation technique, with resemblance to others such as DP-VAE \citep{weggenmann-etal:2022:WWW} or ADePT \citep{krishna-etal-2021-adept}. We choose CusText because it is publicly available\footnote{https://github.com/sai4july/CusText}. DP-VAE lacks available code. The same happens with ADePT, with the addition that its privacy guarantees were later shown to be miscalculated \citep{habernal-2021-differential}. 
   
   CusText replaces tokens to another tokens according to their semantic relevance. The sampling function does not have access to the original similarity scores between original and adversarial tokens. Instead, the scores receive calibrated noise from the exponential mechanism. The strength of the noise is controlled by a parameter $\epsilon$: bigger values for $\epsilon$ mean \textbf{less} noise, and thus the list of candidate tokens tends to be closer to their actual similarities.

    We obfuscated the test set of the Shakespeare dataset with different  $\epsilon$ values and report how the attack performs in Table \ref{tab:custext_as} alongside quality measurements. We evaluate CusText under adversarial attack framework, which is a bit different than the scenario for which this algorithm was designed. CusText was evaluated for text sanitisation, where an obfuscated text should deliver similar results for some task, while using different tokens from the original task. Privacy was assessed not for a downstream classification task, but from unlaballed tasks, such as masked token inference, where an adversary uses a pre-trained BERT model to infer the original tokens since the model is trained with masked language modelling.
    
    CusText generates more adversarial sentences than any attack, but they are in general of poor quality. It is clear that in terms of attack, less noise (bigger $\epsilon$) leads to a weaker attack, but since the sentences are closer to the original ones, similarity measurements are larger. However, both AS and UR are too low when compared to the other attacks, despite the big percentage of perturbed words.

    \begin{table}[h]
        \setlength{\tabcolsep}{1pt}
        \centering
        \footnotesize
        \begin{tabular}{c|cc|ccccccccc}
        \hline
            $\epsilon$ & \textbf{AS} & \textbf{UR} & \textbf{\makecell{Matching \\ POS tags}}  & \textbf{\makecell{Grammar \\ correct}} & \textbf{\makecell{Cosine \\sim.}} & \textbf{\makecell{Levenshtein \\ ratio}} &\textbf{\makecell{\% Perturbed \\ words}} & \textbf{\makecell{Jaccard \\ sim.}}& \textbf{Meteor}& \textbf{\makecell{BERTScore \\ F1}} & \textbf{\makecell{\# Adv. \\ sentences}}\\\hline
            0.5 & .1089& .2539& .0214 & .0379& .0693 & .5998& .6802& .1631&.2311&.8770&3,828\\
            1 & .1112 &.2564 &.0251 &.0390 & .0703 & .6022& .6728& .2208&.2376&.8778&3,822\\
            3 &  .1058 &.2549 & .0318& .0450& .0716& .6232& .6358& .2487&.2742&.8821&3,801\\
            5 & .1134 &.2812 &.0366 &.0515 & .0753 &.6588 &.5700 &.3013 &.3416&.8900&3,766\\
            10 &.1218  &.3275 &.0689 &.0923 &.0827  &.7789&.3663 &.4911 &.5807&.9210&3,456\\ \hline
        \end{tabular}
        \caption{AS and UR for Shakespeare sentences rewritten by CusText, alongside quality measurements.}
        \label{tab:custext_as}
    \end{table}

   \begin{table}[h!]
        \footnotesize
        \begin{tabular}{p{0.47\linewidth} | p{0.47\linewidth}|c}
        \hline
             \textbf{Original} & \textbf{Reformulation} & $\epsilon$  \\
            \hline
            \multirow{2}{*}{god bless you, bottom, god bless you.} & holy bless you, bottom, truth gods you. &0.5 \\ \cline{2-3}
            
            &god pray you, bottom, god bless you.&10\\\hline
            
            no more than my staying here in rome might mean to you in egypt. & reasons most less mind happy where same paris anything without this why part egypt. & 0.5 \\  \cline{2-3}

            &nor more less something staying close on rome might exactly to you in egypt.&10\\\hline
            
            \multirow{2}{*}{what do you mean, do i think so?} & perhaps indeed we mean, way else anything so? &0.5 \\ \cline{2-3}

            &what do else mean, do i thought so?&10\\\hline
            
            \multirow{2}{*}{her name is portia} &her reign only portia&0.5\\ \cline{2-3}

            &her name is portia.&10\\\hline
            
            \multirow{2}{*}{do you intend to walk outside?} & believe really deny for running outside? & 0.5\\  \cline{2-3}

            &anything me punish to walk outside?&10\\\hline
        \end{tabular}
        \caption{Examples of CusText reformulations.}
        \label{tab:custext_samples}
    \end{table}

    We show some examples of reformulations in Table \ref{tab:custext_samples}. It is clear that with a strong amount of noise the sentences can become very different from the original, whereas small amounts of noise may be ineffective to produce any change at all.

\appendixsection{Number of adversarial texts generated by method}
\label{app:num-adv-texts}

\begin{table}[!t]
        \centering
        \footnotesize
        \begin{tabular}{ccc}
        \hline
            \textbf{Dataset} & \textbf{Attack}  & \textbf{\makecell{\# Adversarial \\ sentences}} \\ \hline

            \multirow{9}{*}{TrustPilot\textsubscript{L}}  & IDT & 180 \\
            &Adv. Back-trans.&508\\
            &Back-trans.&505\\
            &TextFooler&393 \\
            &TextBugger&400 \\
            &BAE& 226 \\
            &CusText&998\\
            & Presidio &76 \\
            &ZSTS & 508\\
            \hline

            \multirow{9}{*}{TrustPilot\textsubscript{G}}  & IDT & 922 \\
            &Adv. Back-trans.&3,501\\
            &Back-trans.&3,490\\
            &TextFooler&3,116 \\
            &TextBugger& 2,958\\
            &BAE&  1,959 \\
            &CusText&4,968\\
            & Presidio & 697\\
            &ZSTS &3,501 \\\hline

            \multirow{9}{*}{TrustPilot\textsubscript{A}}  & IDT & 747 \\
            &Adv. Back-trans.&3,735\\
            &Back-trans.&3,715\\
            &TextFooler& 3,456\\
            &TextBugger&3,072 \\
            &BAE& 2,120\\
            &CusText&4,887\\
            & Presidio & 986\\
            &ZSTS & 3,766\\\hline

            \multirow{9}{*}{TOEFL11\textsubscript{S}} & IDT & 2,149 \\
            &Adv. Back-trans.&5,097\\
            &Back-trans.&5,097\\
            &TextFooler& 3,286\\
            &TextBugger&3,212 \\
            &BAE&  2,752 \\
            &CusText&15,942\\
            & Presidio &268 \\
            &ZSTS & 5,097\\\hline
            
            \multirow{4}{*}{TOEFL11\textsubscript{C}}  & IDT & 375 \\
            &TextFooler& 1,301\\
            &TextBugger&1,295 \\
            &BAE&  1,155 \\ \hline
            
            \multirow{9}{*}{Shakespeare} & IDT & 320 \\
            &Adv. Back-trans.&1,008\\
            &Back-trans.&1,324\\
            &TextFooler& 884 \\
            &TextBugger& 643\\
            &BAE& 841\\
            &CusText&3,456\\
            & Presidio &69 \\
            &ZSTS &1,312 \\\hline
            
        \end{tabular}
        \caption{Number of adversarial sentences each algorithm could generate from the set of correctly classified original sentences by the target models.}
        \label{tab:attacks_generated_sentences}
    \end{table}

    The \attack has a tight constraint of ensuring equal POS tags between all adversarial and original tokens. It also finds the words to replace based on their importance regarding two tasks, opposed to one as is the case for the baselines. All of these limit the amount of candidate adversarial sentences \attack can find. We report the total number of adversarial samples generated by each algorithm in Table \ref{tab:attacks_generated_sentences}.

    \attack finds the smallest amount of adversarial sentences across all datasets. The difference to the other algorithms can be smaller, as is the case for TOEFL11\textsubscript{S}, or bigger, as for TOEFL11\textsubscript{C} and TrustPilot\textsubscript{A}.

    CusText consistently is the attack which generates the largest amount of adversarial texts. Interestingly, this two algorithms are in opposite sides when evaluating the effectiveness of their attacks: CusText is in general the poorest performer, and \attack is the strongest.

\section*{Acknowledgements}

This project was undertaken with the assistance of resources and services from the National Computational Infrastructure (NCI), supported by the Australian Government. This project was also supported by the International Macquarie University Research Excellence Scholarship.

The human evaluation section of this study has received ethics approval from Macquarie University (Human Ethics Comm. Approval Code: 5201800393). 

\newpage
\bibliography{main.bib}

\end{document}